%% file: main.tex
\pdfoutput=1
\documentclass[11pt]{article}

\usepackage[final]{acl}

\usepackage{times}
\usepackage{latexsym}

\usepackage[T1]{fontenc}
\usepackage[utf8]{inputenc}

\usepackage{microtype}

\usepackage{inconsolata}

\usepackage{graphicx}

\usepackage{microtype}
\usepackage{hyperref}
\usepackage{url}
\usepackage{booktabs}
\definecolor{darkblue}{rgb}{0, 0, 0.5}
\hypersetup{colorlinks=true, citecolor=darkblue, linkcolor=darkblue, urlcolor=darkblue}

\usepackage{algpseudocode}
\usepackage{cite}
\usepackage{amsmath,amssymb,amsfonts}
\usepackage{graphicx}
\usepackage{textcomp}
\usepackage{xcolor}
\usepackage{listings}
\usepackage{helvet}
\usepackage{courier}
\usepackage{graphicx}
\usepackage{tabularx}
\usepackage{longtable}
\usepackage{amsmath}
\usepackage{subfigure}
\usepackage{xurl}
\usepackage{xspace} 
\usepackage{CJKutf8}
\usepackage{upgreek}
\usepackage{amsthm}
\usepackage{multirow}
\usepackage{amsmath}
\usepackage[ruled,linesnumbered]{algorithm2e}
\usepackage{pgfplots}
\usepackage{tikz}
\usetikzlibrary{patterns}
\pgfplotsset{compat=1.17} 
\usepackage{scalefnt}
\usepackage{floatrow}
\newfloatcommand{capbtabbox}{table}[][\FBwidth]

\usepackage{blindtext}

\newcommand{\eat}[1]{}

\newcommand{\yuxi}[1]{\textcolor{black}{#1}}

\definecolor{color1}{RGB}{123,9,4}
\definecolor{color2}{RGB}{22,53,93}
\definecolor{color3}{RGB}{34,68,122}
\definecolor{color4}{RGB}{141,20,80}
\definecolor{darkred}{rgb}{0.8, 0.0, 0.0}
\definecolor{darkblue}{rgb}{0.0, 0.0, 0.85}

\newcommand{\ourmethod}{\textit{DeTriever}}
\title{DeTriever: Decoder-representation-based Retriever for Improving NL2SQL In-Context Learning}
\author{Yuxi Feng\textsuperscript{1 \textsubscript{*}}, Raymond Li\textsuperscript{1}\thanks{Equal contribution.}, Zhenan Fan\textsuperscript{2}, Giuseppe Carenini\textsuperscript{1}, \\ \textbf{Mohammadreza Pourreza\textsuperscript{3},Weiwei Zhang\textsuperscript{2}, Yong Zhang\textsuperscript{2}}\\
  \textsuperscript{1}The University of British Columbia, Vancouver, Canada\\
  \textsuperscript{2}Huawei Technologies Canada Co., Ltd., Burnaby, Canada\\
  \textsuperscript{3}University of Alberta, Edmonton, Canada\\
  \texttt{\{fyx14, raymondl, carenini\}@cs.ubc.ca,}\\
  \texttt{\{zhenan.fan1, weiwei.zhang2, yong.zhang3\}@huawei.com}, \\
  \texttt{pourreza@ualberta.ca}}

\begin{document}
\maketitle
\begin{abstract}
While in-context Learning (ICL) has proven to be an effective technique to improve the performance of
% on
Large Language Models (LLMs) in a variety of complex tasks, notably in translating natural language questions into Structured Query Language (NL2SQL), the question of how to select the most beneficial demonstration examples remains an open research problem.
% have been proven to be increasingly capable of 
% solving a wide variety of complex tasks, 
% including translating natural language questions into structured query language (NL2SQL). 
% In this work, we focus on select optimizing the selection of demonstration examples 
% A key question in ICL is optimizing the selection of demonstration examples to maximize model performance. 
While prior works often adapted off-the-shelf encoders to retrieve examples dynamically, an inherent discrepancy exists in the representational capacities between the external retrievers and the LLMs. Further, optimizing the selection of examples is a non-trivial task, since there are no straightforward methods to assess the relative benefits of examples without performing pairwise inference.
% Motivated by the findings that hidden states of LLMs contain richer semantic information
To address these shortcomings, we propose \ourmethod, a novel demonstration retrieval framework that learns a weighted combination of LLM hidden states, where rich semantic information is encoded. To train the model, we propose a proxy score that estimates the relative benefits of examples based on the similarities between output queries. Experiments on two popular NL2SQL benchmarks demonstrate that our method significantly outperforms the state-of-the-art baselines on one-shot NL2SQL tasks.
\end{abstract}

% While in-context Learning (ICL) has proven to be an effective technique to improve the performance of Large Language Models (LLMs) in a variety of complex tasks, notably in translating natural language questions into Structured Query Language (NL2SQL), the question of how to select the most beneficial demonstration examples remains an open research problem. While prior works often adapted off-the-shelf encoders to retrieve examples dynamically, an inherent discrepancy exists in the representational capacities between the external retrievers and the LLMs. Further, optimizing the selection of examples is a non-trivial task, since there are no straightforward methods to assess the relative benefits of examples without performing pairwise inference. To address these shortcomings, we propose Detriever, a novel demonstration retrieval framework that learns a weighted combination of LLM hidden states, where rich semantic information is encoded. To train the model, we propose a proxy score that estimates the relative benefits of examples based on the similarities between output queries. Experiments on two popular NL2SQL benchmarks demonstrate that our method significantly outperforms the state-of-the-art baselines on one-shot NL2SQL tasks.

\input{sec/introduction}
\input{sec/relatedwork}

\input{sec/method}
\input{sec/experiment}
\section{Conclusion}
In this work, we propose \ourmethod, an approach to retrieve demonstration examples for ICL, where a weighted combination of LLM layer transformations is used to dynamically retrieve demonstration examples from the training set. We first perform an in-depth study to determine the proxy metric that best approximates the relative benefits of demonstration examples and then convert the proxy metric as the target for the contrastive loss objective used to fine-tune \ourmethod. Our experiments on two popular benchmarks demonstrated the effectiveness of our approach. While we focus on NL2SQL in this work, our method can be naturally applied to other open-ended generation tasks such as code completion and question answering. For future work, we plan to
% adapt \ourmethod~for other domains while conducting
conduct further experiments to analyze the performance in multi-shot ICL.
\newpage

\section{Limitations}
The publicly available datasets used in this work are limited to English. Datasets in other languages (e.g., Vietnamese, Swahili) require LLMs with sufficient capabilities to understand the natural language questions containing domain-specific jargon. Further, while the scope of our experiments is limited to NL2SQL, we recognize the generalizability of our proposed method to other downstream tasks (e.g., question answering, code generation, etc.). However, one important assumption is the correlation between the similarity of the ground-truth answer (estimated through LLM embeddings) and the relative benefits of the corresponding examples for in-context learning demonstrations. This assumption is influenced by several task-specific factors, such as the complexity of the problem and the nature of the output space, and requires analysis to verify for any given task. Lastly, we only evaluate the performance on the NL2SQL benchmarks based on execution accuracy. Execution accuracy is not a perfect metric since it does not directly measure logical equivalence 
and is dependent on the correctness of the underlying database and the ability of the SQL queries to retrieve the expected results. Lastly, there may be other criteria such as query efficiency and simplicity, that could be important for the end-user.

\bibliography{custom}

% \appendix

% \section{Example Appendix}
% \label{sec:appendix}

% This is an appendix.

\end{document}

%% file: sec/introduction.tex
\section{Introduction}
\label{sec:intro}
% Large language models (LLMs) have impressively enhanced sample efficiency and performance across numerous natural language processing (NLP) tasks.
As large language models (LLMs) have become increasingly capable of solving a wide variety of complex tasks \citep{brown2020language, touvron2023llama1, touvron2023llama, rozière2024codellama}, they have become an integral component in many user-facing applications. 
In particular, the task of translating natural language questions into structured query language (NL2SQL) in the context of existing databases, has received a lot of attention due to its relevance in enhancing business intelligence tools and offering significant potential to make data analytics and exploration more accessible to non-technical users.
% A notable area of improvement is in translating natural language questions into SQL (NL2SQL) queries. 
While the recent development of LLMs pre-trained in code completion tasks \citep{rozière2024codellama} has offered a significant boost in NL2SQL capabilities, the current performance of these models still falls short of the standards required for deployment in production environments.
This is mainly due to the many challenges including dealing with domain-specific jargon along with constructing complex queries, which requires an advanced understanding of relationships between database schemas and the underlying intentions of natural language questions \citep{deng-etal-2022-recent, swamidorai2023translating}.

\begin{figure*}
    \centering
    \includegraphics[width=\textwidth]{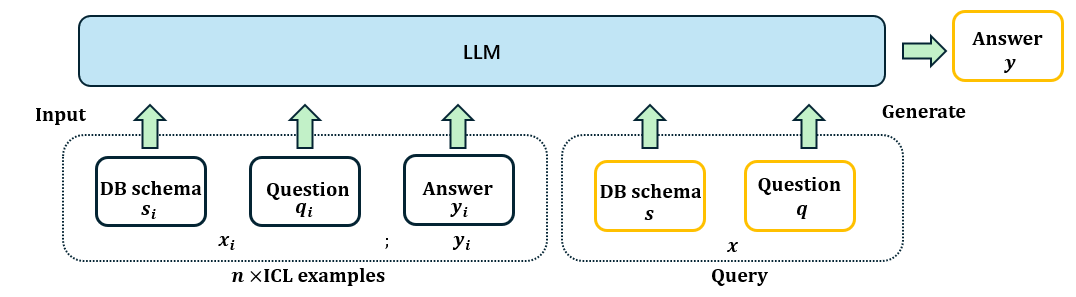}
    \caption{Examples of in-context learning (ICL) for NL2SQL.}
    \label{fig:icl_eg}
\end{figure*}

Research on LLMs has unveiled the emergent abilities \citep{wei2022emergent} of models such as GPT3 \citep{brown2020language} and Llama \citep{touvron2023llama1, touvron2023llama} to improve task performance through the use of in-context learning (ICL), where demonstration examples are integrated into the input prompt during inference to allow the model to make use of aspects from the demonstrations \citep{min-etal-2022-rethinking}. As proven in a variety of complex tasks such as mathematical reasoning \citep{wei2022cot}, ICL has become a cost-efficient alternative to supervised fine-tuning which requires significant computational resources to update the model parameters. In the domain of NL2SQL, recent studies \citep{pourreza2023din, gao2023texttosql} have adopted ICL to achieve state-of-the-art performance on popular NL2SQL benchmarks including Spider~\citep{yu2018spider} and BIRD~\citep{li2024bird}. This is done by prepending labeled examples to the task prompt, allowing the model to draw an analogy from the mapping between the problem description and the corresponding query. \yuxi{\autoref{fig:icl_eg} shows the overview of model inference with ICL demonstrations on the NL2SQL task. In our setup, demonstrations of (database (DB) schema, question, answer) triples are prepended to the NL2SQL question.}
% models leveraging in-context learning, which involves providing a few example questions and SQL pairs in the prompt, 
% have achieved state-of-the-art performance \citep{pourreza2023din, gao2023texttosql} on NL2SQL benchmarks: Spider\citep{yu2018spider} and BIRD\citep{li2024bird}. Recently large language models (LLMs) like GPT4 \citep{openai2023gpt4}, Llama-2 \citep{touvron2023llama}, and Code Llama\citep{rozière2024codellama} have raised people's attention with their emergent ability \citep{wei2022emergent} to learn from a few examples in the context, which is so-called in-context learning (ICL). The key idea of in-context learning is to learn from analogy to examples in a given prompt. Different from supervised learning requiring a training stage that uses backward gradients to update model parameters, ICL does not conduct parameter updates and directly performs predictions on the pre-trained language models. Few-shot ICL of LLMs have been proven to achieve comparable performance with supervised fine-tuning on smaller models on numerous tasks like mathematical reasoning \citep{wei2022cot}. 

Since model performance can vary widely depending on the choice of in-context examples \citep{liu-etal-2022-makes},
a key challenge is determining the most effective strategy to select demonstration examples for any given task. 
% which examples can best benefit a given task.
While prior studies have examined the usefulness of demonstrations based on characteristics of the examples \citep{wei2022emergent}, a more effective strategy is to dynamically retrieve demonstrations for a given example during inference. Following recent works on informational retrieval (IR), many existing proposals \citep{das-etal-2021-case, khattab2022demonstrate} have utilized off-the-shelf encoder models such as Sentence-BERT \citep{reimers-gurevych-2019-sentence} and Contriver \citep{izacard2021contriever} to retrieve ICL examples based on the embeddings similarity between the task input prompt, while more recent studies \citep{rubin-etal-2022-learning, shi2023replug} have opted to fine-tune the retriever based on the LLM perplexity of the target output. However, this fine-tuning strategy cannot be trivially adapted for code completion tasks such as NL2SQL
%since there could be an arbitrary number of prediction that solves the task while the discrete signal of completion accuracy cannot effectively differentiate the relative benefits of individual demonstrations.
since there is an arbitrary number of correct predictions (i.e., multiple different queries retrieving the same records) for each question,
% there could be multiple prompts that make the LLMs generate correct answer for a given input, 
while the discrete signal of inference accuracy cannot effectively differentiate the relative benefits between individual ICL examples. 
Further, there exists a fundamental gap between the retrieval encoders and the LLM due to the inherently more sophisticated representational capabilities of the LLM to differentiate the intricate nuances embedded within the task prompt. Lastly, as evident by prior probing studies measuring the differences between information captured at different locations of the model \citep{chuang2023dola}, it remains a non-trivial task to effectively leverage LLM hidden states as representations for a given prompt.
% This is evident from probing studies on 

% One key question is how to retrieve the examples to input as a prompt to LLMs. Traditional information retrieval (IR) methods train an encoder (e.g., Sentence-BERT \citep{reimers-gurevych-2019-sentence} and Contriever \citep{izacard2021contriever}) to compute examples embeddings. Recently people have started to leverage LLM's own hidden information to compute text embedding. \citet{shi2023replug} treat LLM as a black box to retrieve relevant documents from an external corpus using an off-the-shelf retrieval model, where retrieved documents are prepended to the input context and fed into the black-box LLM.  \citet{wang2024improving} proposes to use the last hidden layer of end-of-sentence (EOS) token as the sentence embedding.

\begin{figure*}[ht!]
    \centering
    \includegraphics[width=\textwidth]{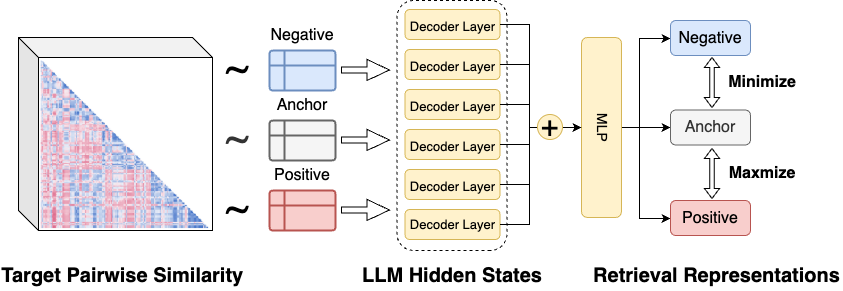} % 
    \caption{Overview of the training process of our proposed \ourmethod~method.}
    \label{fig:detriever} 
\end{figure*}

To overcome the above-mentioned challenges, we propose \ourmethod, a novel ICL retrieval framework that learns  a weighted combination of LLM layer transformations to dynamically retrieve examples based on similarities. To train this model, we first perform an in-depth study to determine the proxy metric that best approximates the relative benefits of demonstration examples. Based on the findings from our analysis, we convert the proxy metric as the target for the contrastive loss used to fine-tune \ourmethod. Our experiments on two popular benchmarks demonstrated the effectiveness of our approach where we achieved significant improvements over state-of-the-art baselines.
The contributions of our work can be summarized as the following:
% We summarize our contributions as follows:

\begin{itemize}
    \item We propose \ourmethod, a novel approach that learns a weighted combination of LLM's layer transformations as representation for retrieval demonstration examples.
    \item We train \ourmethod~by selecting a proxy metric for measuring the relative benefit between examples for ICL and using it as the target for supervised contrastive loss. 
    \item We perform comprehensive experiments on two popular benchmarks to demonstrate the effectiveness of our approach, where \ourmethod~significantly outperforms state-of-the-art baselines for NL2SQL.
    \item We conduct a detailed analysis of our approach and reveal important insights for developing future methods to retrieve demonstration examples for practical applications.
\end{itemize}

%% file: sec/relatedwork.tex
\section{Related Work}
\label{sec:relatedwork}

\paragraph{NL2SQL}
Traditionally, NL2SQL tasks are solved by sequence-to-sequence encoder-decoder models \citep{choi-etal-2021-ryansql}. Recently, large language models (LLMs) have impressively enhanced sample efficiency and performance across numerous natural language processing (NLP) tasks. A notable area of improvement is in translating natural language questions into SQL (NL2SQL) queries. In this domain, models leveraging in-context learning, which involves providing a few example questions and SQL pairs in the prompt, have achieved state-of-the-art performance \citep{pourreza2024sqlencoder,pourreza2023din, gao2023texttosql} on the largest existing comprehensive 
and cross-domain NL2SQL benchmarks: Spider\citep{yu2018spider} and BIRD\citep{li2024bird}. 

\paragraph{LLMs and Retrieval-augmented ICL} Recently large language models (LLMs) like GPT4 \citep{openai2023gpt4}, Llama-2 \citep{touvron2023llama}, and Code Llama\citep{rozière2024codellama} have raised people's attention with their emergent ability \citep{wei2022emergent} to learn from a few examples in the context, which is so-called in-context learning (ICL). The key idea of in-context learning is to learn from analogy to examples in a given prompt. Different from supervised learning requiring a training stage that uses backward gradients to update model parameters, ICL does not conduct parameter updates and directly performs predictions on the pre-trained language models. Few-shot ICL of LLMs have been proven to achieve comparable performance with supervised fine-tuning on smaller models on numerous tasks like mathematical reasoning \citep{wei2022cot}. 
One key question is how to retrieve the best examples to input as a prompt to LLMs for a given problem. 

Related work of in-context retrieval can be organized into two categories. (1) Use an external encoder to retrieve examples, e.g., Sentence-BERT (S-BERT) \citep{reimers-gurevych-2019-sentence} or Dense Passage Retriever (DPR) \citep{karpukhin-etal-2020-dense}.
(2) Retrieval-augmented language models 
\citep{izacard2021contriever, JMLR:v24:23-0037, borgeaud2022improving, ma2023finetuning}
% (e.g., \citet{borgeaud2022improving}, \citet{}, \citet{ma2023finetuning}). 
These works are based on the encoder-decoder transformer architectures, where the model representation is trained to retrieve examples based on its hidden representations.
% Retrieved examples are embedded in theoften encoded to the same latent space as decoder hidden layers \citep{}.  where the ICL examples are embedded in the same latent space as decoder hidden layers. 
As a result, applying such an approach requires further fine-tuning to recover from the modified representation, and cannot be directly applied for decoder-only architectures.
% thus cannot be directly applied in NL2SQL tasks.

Recently people have started to leverage LLM's own hidden information to compute text embedding. \citet{shi2023replug} treat LLM as a black box to retrieve relevant documents from an external corpus using an off-the-shelf retrieval model, where retrieved documents are prepended to the input context and fed into the black-box LLM.  \citet{wang2024improving} proposes to use the last hidden layer of the end-of-sentence (EOS) token as the sentence embedding for retrieving ICL examples.

Unlike all previously mentioned methods which either use an external encoder-based retriever or simply use the last hidden layer of the EOS token, we propose to use the internal hidden states of LLMs to retrieve in-context examples.

%% file: sec/method.tex
\section{Approach}
\label{sec:method}

\subsection{Problem Formalization}
Given a database schema $s$ and the natural language question $q$, the goal of the NL2SQL task is to generate the SQL query $y$ that correctly retrieves database records to answer the question according to the given schema. We leverage a large language model $M$ for the query generation task, where the database schema and question are formatted into an input prompt $x = [s; q]$ such that query prediction $\hat{y} = M(x)$. 
For ICL inference, we prepend demonstration examples to the input prompt such as $\hat{y} = M([x_1, y_1; \dots; x_n, y_n; x])$, where question-answer pairs $(x_i, y_i)$ are sampled from a labeled training set $\mathcal{D}_{\textrm{train}}$. 
% we assume a training corpus $\mathcal{D}$ with input-target pairs, where  where the examples $P^* \subset D = [P_i, Q_i; \dots; P_n, Q_n]$ are prepended to the problem $Q$, we focus on finding useful examples from a corpus $\mathcal{D}$ that helps to solve the problem $P$. 
% Formally, given a corpus of schema-question-SQL triplets $\mathcal{D} = \{S_i, N_i, Q_i\}_n$, 
For demonstration retrieval, we first apply a retrieval model $\mathcal{R}$ to compute the retrieval representation for the input of all examples in $\mathcal{D}_{\textrm{train}}$. During inference for problem description $x$, we select an ICL example $I(x)$ based on the highest dot-product similarity between the retrieval model representations.
\begin{equation}
\label{eq:retrieval-formalism}
I(x) = (x_i, y_i) = \underset{(x_i, y_i) \in \mathcal{D}_{\textrm{train}}}{\textrm{argmax}} \mathcal{R}(x)\cdot \mathcal{R}(x_i)
\end{equation}

\input{fig/layer_result}

% retrieval model $\mathcal{R}$ to the  prompt $x$ and and all examples  returns $k$ examples $\mathcal{R}(x)=\{x_i, y_i\}_{i=1}^k\subset \mathcal{D}_{\textrm{train}} $ based on the similarity of the representations.
Since the probability of predicting the correct query can be measured through execution accuracy by comparing the predicted query $\hat{y}=M([I(x); x])$ against the ground-truth query $y$ over an evaluation corpus $\mathcal{D}_{\textrm{test}}$, where $\textrm{Acc}(x, \hat{y}, y) \in \{0, 1\}$, the objective of optimizing the retriever $\mathcal{R}$ can be expressed in \autoref{eq:problem-formalism}.

\begin{equation}
    \begin{aligned}
        \label{eq:problem-formalism}
        \underset{{\mathcal{R}}}{\textrm{argmax}} &\sum_{(x_j, y_j) \in \mathcal{D}_{\textrm{test}}}\textrm{Acc}(x_j, M([I(x_j); x_j]), y_j)\quad \\ 
        &\textrm{s.t.}~I(x_j) \in \mathcal{D}_{\textrm{train}}
    \end{aligned}
\end{equation}

% the performance on 
% in-context-learning to improve the performance n order to improve the generation performance of LLM $Q = M(S^*, N^*, Q^*,S,N)$, where $(S^*, N^*, Q^*)\in\mathcal{D}$ is selected by a retriever $R$. The retriever $R$ is a function that takes as input a pair of the database schema and NL question $(S,N)$ and a corpus $\mathcal{D}$ and returns the best one-shot prompt $(S^*, N^*, Q^*)$. At run-time, the retriever $R$ computes the embedding of $(S,N)$ pair $E(S,N)\in \mathbb{R}^d$ and retrieves the prompt of which embedding are the closest to the question pair embedding. 

% \subsection{Input format}

% \subsection{Decoder-based Embeddings}
\subsection{LLM Hidden States}
\label{sec:decoder-based}
Although prior studies have often employed external encoders, such as Sentence-BERT~\citep{reimers-gurevych-2019-sentence} and DPR~\citep{karpukhin-etal-2020-dense}, to retrieve in-context exemplars for the LLM,  there is a fundamental gap between these models and the LLM. This is because the LLM has more advanced abilities to capture semantic meanings and encode the subtle nuances of the input. Therefore, we directly use the hidden states of the LLM as input embeddings to the retriever. However, unlike encoder models such as BERT \citep{devlin-etal-2019-bert}, where the hidden states of the last layer can be directly used as representations of the sequence, there are no straightforward methods of adapting an autoregressive LLM for encoding the input problem for retrieval. While it may be tempting to simply use the last layer as the problem embedding \citet{wang2024improving}, hidden states of lower layers can have better representation in the semantic meaning of the tokens \citep{chuang2023dola} which may lead to improvements in retrieval qualities.

To test this hypothesis, we perform an analysis on the performance of each layer for retrieving in-context examples. Specifically, for layer $\ell$ of the LLM, we use the mean-pooled or end-of-sequence (EOS) token hidden state of the problem prompt $x = [s; q]$ to retrieve the most similar examples in the training set $\mathcal{D}_{\textrm{train}}$ based on cosine similarity \yuxi{of the $\ell$-th layer hidden-states $M_{\ell}$ of the large language model}. (\autoref{eq:layer-retrieval}).

\begin{equation}
\vspace{-1em}
\label{eq:layer-retrieval}
    \mathcal{I}(x) = \underset{(x_i, y_i) \in \mathcal{D}_{\textrm{train}}}{\textrm{argmax}} M_{\ell}(x) \cdot M_{\ell}(x_i)
\end{equation}
\vspace{1em}

As illustrated in \autoref{fig:layer_result}, we see that the hidden representations from lower-mid layers (10-20) have a significantly better execution accuracy compared to the last layer (40). Motivated by this finding, we propose to learn a weighted combination over the layers \citep{Bahdanau2015NeuralMT} to effectively leverage the different granularity of information encoded in each layer of the LLM, since we do not know which layer performs the best beforehand. Specifically, given the hidden states representation for question $x$ in each LLM layer $\{h_\ell(x)\}_{\ell=0}^L$ where $L$ is the number of layers of LLM, we first use a 3-layer multi-layer-perception (MLP) \citep{MURTAGH1991183} to transform the hidden states of each layer before computing the final retrieval representation $\mathcal{R}(x)$ as the weighted sum of layer transformations. 
% by learning a weight parameter $w_\ell \in \mathbb{R}$ for each layer.

\begin{align}
    \mathcal{R}(x)&=\sum_{\ell=1}^Lw_\ell\cdot \text{MLP}_\ell(h_\ell(x))
\end{align}

% Unlike traditional S-BERT \citep{reimers-gurevych-2019-sentence} and DPR \citep{karpukhin-etal-2020-dense} who uses an additional encoder to compute text embeddings, we directly use the hidden state of the LLM to compute text embeddings since internal hidden states of LLMs contain nuanced information for supporting its decision \citep{azaria-mitchell-2023-internal}.
% Unlike \citet{wang2024improving} who directly leverage the last hidden state of the EOS token as embedding, we argue that simply use one specific layer of one specific token is arbitrary and might lead to a suboptimal result. As discussed in \citet{chuang2023dola}, lower hidden layer of a decoder can better represent the semantic meaning of the token and may improve the quality of retrieval. As shown in Fig.~\ref{fig:layer_result}, representation in mid hidden layers (layer 10-20) can achieve a significantly better execution accuracy then the last hidden layer. In order to leverage the information in each hidden layer of the LLM, we propose to learn an attention weight (cite?) of the hidden states. 

\subsection{Answer similarity-based training objective}
\label{sec:exeaccu}
To optimize \autoref{eq:problem-formalism}, we need a labeled training set that indicates whether each in-context example $(x_i, y_i) \in \mathcal{D}_{\textrm{train}}$ is beneficial for the question $x$.
However, since execution accuracy only provides binary signals, there are no trivial methods for directly computing the relative improvement of using an in-context example for any question $x$ without performing pairwise inference. Further, there are often cases when a question can be correctly predicted for nearly all demonstration examples in $\mathcal{D}_{\textrm{train}}$ (and vice versa), which reduces the number of usable training examples.
Therefore, we propose to use an approximate metric $\textrm{Sim}([x_i; y_i], [x; y])$ to approximate the benefits of using $[x_i; y_i]$ as examples to predict $\hat{y}$.

However, during model inference, since the target query is unknown, $\textrm{Sim}([x_i; y_i], [x; y])$ is not directly computed.
Instead, we propose an indirect approach to predict $\textrm{Sim}([x_i; y_i], [x; y])$ by training the retrieval model $\mathcal{R}$ using supervised contrastive loss. The main intuition is to align the retriever embedding space (problem description as inputs)
% where the inputs are NL2SQL problem descriptions (schema + natural language question) 
with the similarity between the concatenated representation of problem description and ground-truth SQL query. This contrasts the prior work by \citet{nan2023enhancing}, where a proxy query is first generated through zero-shot inference, resulting in significant added latency. In practice, we select $n_\textrm{pos}$ positive and $n_\textrm{neg}$ negative examples for a given anchor $x$ based on the dot-product of LLM hidden state representations. We formally define the contrastive loss in \autoref{eq:loss}, where $\tau$ is a constant temperature.

% One key question to train a text retriever is the design of the objective. We propose to select \yuxi{in-batch} positive or negative examples by the similarity of answer representations. In our settings, we pass the example $(x_i, y_i)$ to LLM decoder and use the hidden state representation as contextualized answer representation. Given \yuxi{an anchor example} $(P, Q)$, we sample $n_\text{pos}$ positive examples $\mathcal{P}_{\text{pos}}$ and $n_\text{neg}$ negative examples $\mathcal{P}_{\text{neg}}$ based on the cosine similarity of answer representations $\textrm{Sim}([P^*; Q^*], [P; Q])$. Then we train the retriever with the following supervised contrastive loss function:

%For one specific input tuples $(S^*, N^*, Q^*,S,N)$ where $(S^*, N^*, Q^*)\in \mathcal{D}$, it is considered as a positive example if the final output $Q = M(S^*, N^*, Q^*,S,N)$ can be executed accurately, otherwise it is a negative example. Then we can train the retriever by Triplet Ranking Loss as follows.

\begin{equation}
    \begin{aligned}
        &L = 
        % \frac{-1}{n_{\text{pos}}}
        \sum_{x\in\mathcal{D}}\sum_{x_i\in \mathcal{P}_{\text{pos}}}\log 
        \frac{\text{exp}(\mathcal{R}(x)\cdot \mathcal{R}(x_i)/ \tau) }{\sum_{x_j\in 
        % \mathcal{P}_{\text{pos}} \cup \mathcal{P}_{\text{neg}}
        \mathcal{P}} \text{exp}(\mathcal{R}(x)\cdot \mathcal{R}(x_j)/ \tau)  }
    \end{aligned}
    \label{eq:loss}
\end{equation}

%% file: fig/layer_result.tex
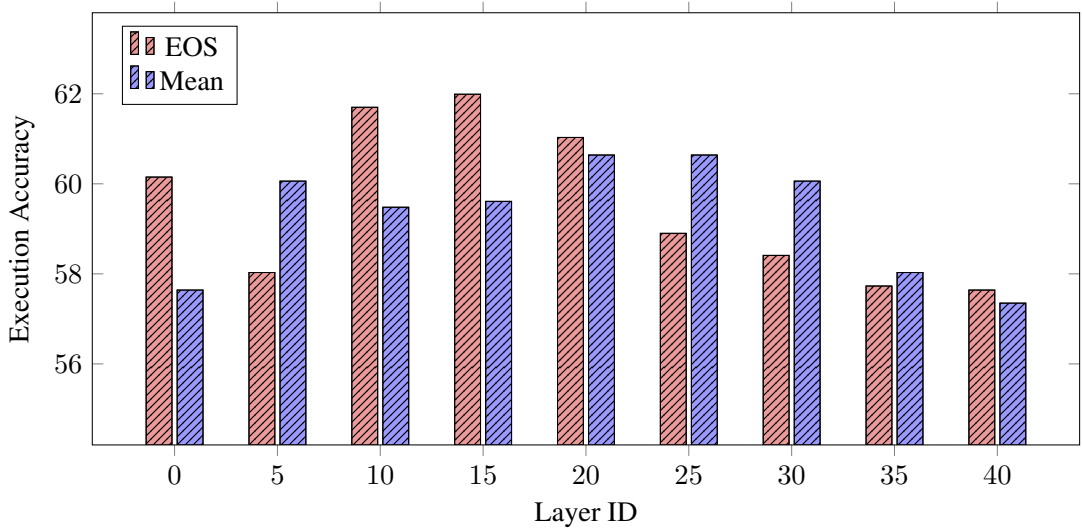
\begin{figure*}[htb] %插入图片

\centering %图片居中
\resizebox{0.9\columnwidth}{!}{
\begin{tikzpicture}
\begin{axis}[
ybar,
legend pos=north west,
xlabel=Layer ID, %x坐标名
ylabel=Execution Accuracy, %y坐标名
ymin=55,
ymax=63,
enlargelimits=0.1,
width=15cm, height=7.5cm,
%x tick label style={shift={(axis cs:0.5,0)},anchor=east,rotate=60,},
%symbolic x coords={\text{[0,20]}, \text{[20,40]},\text{[40,60]},\text{[60,80]},\text{[80,100]}},
]
    \addplot [ybar,fill=darkred!40!white,draw=black, 
    postaction={pattern=north east lines} ] coordinates{
    %\addplot+[ybar,fill=RYB1] plot coordinates {
    (0,60.15)
    (5,58.03)
    (10,61.70)
    (15,61.99)
    (20,61.03)
    (25,58.90)
    (30,58.41)
    (35,57.73)
    (40, 57.64)
    };
    \addlegendentry{EOS}

    \addplot [ybar,fill=blue!40!white,draw=black, 
    postaction={pattern=north east lines} ] coordinates{
    %\addplot+[ybar,fill=RYB1] plot coordinates {
    (0,57.64)
    (5,60.06)
    (10,59.48)
    (15,59.61)
    (20,60.64)
    (25,60.64)
    (30,60.06)
    (35,58.03)
    (40, 57.35)
    };
    \addlegendentry{Mean}
\end{axis}
\end{tikzpicture}
}
\caption{One-shot ICL results (execution accuracy on development set of Spider dataset) of CodeLlama-13b where prompt is retrieved by representations computed in each decoder layer. Here \emph{Mean} represents average hidden state of a specific layer, where \emph{EOS} represents the hidden state of EOS token in a specific layer. The one-shot Prompt is retrieved by the one with biggest cosine similarity.} % 设置caption
\label{fig:layer_result}  % 设置用于reference的label
\end{figure*}

%% file: sec/experiment.tex
\section{Experiments}
\label{sec:experiment}

\begin{table*}[tbp]
\small
 \centering
 \begin{tabular}{lcccc}\toprule
    & \multicolumn{2}{c}{Spider} & \multicolumn{2}{c}{BIRD}
    \\
    \cmidrule(lr){2-3}\cmidrule(lr){4-5}
             & Exe Acc. $\uparrow$  & Error Rate $\downarrow$ &  Exe Acc. $\uparrow$  & Error Rate $\downarrow$ \\\midrule
    Zero-shot &56.74&16.3 &26.66&42.75\\
    Random 1-shot &56.80&16.0&20.79&45.43\\
    \midrule
    \emph{External embedding models}\\
    Contriever&56.83&16.1&20.51&46.83 \\
    OpenAI Embeddings& 57.06&14.3&20.99&45.38\\
    \midrule
    \emph{LLM representations}\\
    Mean~(best)& 57.90&15.5&28.55&41.92\\
    EOS~(last) &57.64&14.4&27.77&42.05\\
    EOS~(best) &61.99 &11.2&29.07&42.96\\
    \midrule
    \ourmethod  & \textbf{68.38}&\textbf{9.77}& \textbf{29.73} & \textbf{38.59}\\
    \bottomrule
 \end{tabular}
 \caption{Results on Spider dataset and BIRD dataset.}
 \label{tab:mainresult}
\end{table*}

\subsection{Datasets}
Our evaluation involved two comprehensive cross-domain datasets, Spider \citep{yu2018spider} and BIRD \citep{li2024bird}. Spider contains 10,181 questions linked to 5,693 SQL queries across 138 domains and 200 databases, divided into 8,659 training, 1,034 development, and 2,147 test examples from unique databases. BIRD comprises 12,751 question-SQL pairs from 95 databases across 37+ domains, including blockchain and healthcare, totaling 33.4 GB. The dataset breakdown for BIRD includes 1,534 development, 9,428 training, and 1,789 test queries. 
Following prior works \citep{pourreza2023din}, the development sets from both datasets are used for evaluating the models without performing hyperparameter tuning. 

Different from prior works \citep{pourreza2024sqlencoder} which use in-domain (ID) examples they assume references have the same DB schema as the query, we consider a more practical out-of-domain (OOD)\footnote{Note that this OOD is different from the traditional setting of OOD where a model is trained in data in one domain and tested on another domain.} setting in the following experiments, where the DB schema of the query is different from it of all references. The OOD scenario reflects the practical challenges in real-world applications, where it may be uncommon to have accessible labeled examples within the same database schema. 
% For  the models  from both datasets for evaluation without performing hyper-parameter tuning. We

%For each question, we sampled 20 others (or the total number available per database if fewer than 20) sample questions from the same database and calculated the mean of all three similarity metric for the pair as the prediction label. 

\subsection{Experimental Settings}
We utilize the training set of BIRD and Spider respectively to train the retriever. We use Code-Llama-13b \citep{rozière2024codellama} as the large language model to generate SQL queries. Code-Llama-13b contains 41 layers of transformer blocks \citep{Vaswani+2017}. In order to reduce the parameter of the retriever, we only consider the hidden states for every 5 layers (e.g., layer 0, layer 5, ..., layer 40). The hidden layer shape in the 3-layer MLP is 1024, and the learned embedding dimension is 512.

In training, we train the retriever with 10k steps and store the checkpoints every 1k steps. The training takes around 10 minutes in a single NVIDIA V100 GPU node. We tuned all hyperparameters on Spider held-out development set and applied the best hyperparameter setting in Bird dataset. In the training phase, we set the number of negative examples $n_{\text{neg}}=100$ and tuned the number of positive examples $n_{\text{pos}}\in\{5,10,20,40,60,80\}$ and finally choose $n_{\text{pos}}=40$. We also tuned batch size $\in \{16,32,64\}$ and finally chose the batch size to be 64. We use AdamW \citep{Loshchilov2019DecoupledWD} as an optimizer. The learning rate is $1e-4$, weight decay is $0.01$, beta1 is 0.9, and beta2 is 0.98, following the default setting. The temperature parameter for supervised contrastive loss $\tau=0.07$ by default. 

%We apply linear warmup to the optimizer, and the warmup ratio is 0.06.

In the inference phase, we first retrieve the example with the largest cosine similarity by our trained retriever and ask the large language model (CodeLlama-13b) to generate SQL queries given the retrieved example as a prompt. Following previous works \citep{pourreza2023din}, we use greedy decoding to generate the most confident tokens. 

The evaluation metric is the execution accuracy (Exe Acc.) of predicted SQL queries. We also include the percentage of generated SQLs that cannot be successfully compiled (Error Rate) for additional information.

\subsection{Baselines}
We tested the following methods for one-shot in-context-learning on the NL2SQL task.
\begin{itemize}
    \item \textbf{Random}: Randomly select example as a prompt without any retrieval.
    \item \textbf{External embedding models}: (1) \textbf{Contriever} \citep{izacard2021contriever}: an unsupervised encoder-based retriever using contrastive learning. (2) \textbf{OpenAI Embeddings}\footnote{\url{https://platform.openai.com/docs/guides/embeddings/what-are-embeddings}}: A general text embedding provided by OpenAI company. We choose to use \emph{text-embedding-3-large} to compute text embedding for text retrieval. 
    \item \textbf{LLM representations}: Use the mean-pooled or EOS hidden states of a specific layer of LLM to compute the similarity. (1) \textbf{Mean(best)}: the result using mean-pooled hidden representation from the best LLM layer. (2) \textbf{EOS(last)}\citep{wang2024improving}: the result using EOS token representation from the last hidden layer. (3) \textbf{EOS(best)}: the result using EOS token representation from the best LLM layer.
\end{itemize}

%(Our proposed attention-based method)

\subsection{Results}
% We report the results from our experiments in \autoref{tab:mainresult}. 
From the results reported in \autoref{tab:mainresult}, we see that our approach achieved the best performance across both benchmarks, with a $11.58$ point increase in execution accuracy over the random one-shot baseline on Spider and $8.94$ on Bird. Furthermore, we are able to validate our hypothesis that LLM hidden states offer a more effective means of encoding task descriptions for demonstration retrieval, where the representations derived from mean-pooling and end-of-sequence (EOS) hidden states surpassed the performance of both Contriever and the state-of-the-art OpenAI embeddings. It is also worth noting that the EOS token hidden states from the best layer demonstrated a $3.35$ and $1.30$ improvements in execution accuracy over the last layer.  This observation challenges previous studies' reliance on using the final layer representations for retrieval, suggesting a reevaluation of the presumed equivalence to the hidden states of encoder-only models. Lastly, we are pleased to report that \ourmethod outperforms both \textit{EOS (best)} and \textit{Mean (best)}, as they represent the upper-bound performance for retrieval using single-layer representations. This further confirms our hypothesis that using a combination of hidden layers can more effectively leverage the different granularity of information encoded in each layer of the LLM.

\subsection{Analysis}

\subsubsection{Training Target}
In the first analysis, we study the effects of using different proxy scores as approximations for the relative benefits of demonstration examples. Specifically, we select the pairwise embedding similarity of different representations as the target for the contrastive loss presented in \autoref{eq:loss}.
In the results presented in \autoref{tab:objective-analysis}, we compare the performance of mean-pooled and EOS representation of using only the target SQL query (Query-only) with using both the problem description and the query (Problem+Query). We also include the results for retrieving examples based on the problem description (Problem-only) as a reference for comparison. Since using the problem description does not require knowledge of the ground-truth query, we simply use the best-performing layer from \autoref{tab:mainresult} without the need for training.

\begin{table}[htbp]
\small
 \centering
 \begin{tabular}{lcc}\toprule
             & Exe Acc. $\uparrow$ & Error Rate $\downarrow$ \\\midrule
    Problem-only (no training) & 62.0 &11.2 \\
    \hline
    Query-only Mean & 67.8& 8.9\\
    Query-only EOS & 67.4 & 10.0 \\
    Problem+Query Mean & 62.4 & 12.7\\
    Problem+Query EOS & 68.4 & 9.8\\
    \bottomrule
 \end{tabular}
 \caption{Results on Spider dataset with different targets.}
 \label{tab:objective-analysis}
\end{table}

From the experiments, we see that training the retriever with Query-only does not significantly reduce the task performance, causing a less than $1$ point accuracy decline over the EOS of Problem+Query. 
This finding stands in contrast to the results reported by \citet{min-etal-2022-rethinking}, where they find that using only the task description without the ground-truth label can achieve similar performance compared to using input-label pairs. Since they only performed experiments on classification and multi-choice tasks, we hypothesize that for code-completion tasks such as NL2SQL, where there can be high variations in potential predictions, the model will often resort to copying or augmenting the target from the demonstration examples rather than relying on the priors from pre-training. This tendency renders the model particularly sensitive to the choice of demonstration examples, highlighting the importance of an efficient ICL retriever for optimal performance. This hypothesis also explains the poor performance of using the mean-pooled Problem+Query representations as training target, since problem description dilutes the information from the target query due to the autoregressive nature of the decoder-only LLM.

% 1. Target: answer eos, question-answer mean, ... 
% each layer

% 2. Layer of input to use 
% Not find any difference.

% \yuxi{(Yuxicomment: there might be some bugs, I just found out we directly loaded hidden states from pt, however  the hidden states should be different when the embedding changes.)}

% 2. Layer of input to use

\subsubsection{Number of Positive Anchors}

\input{fig/num_pos}

We also study the effects of classifying different numbers of the most similar examples as positives in the contrastive objective. From the results presented in \autoref{fig:num_pos}, we see that the performance increases along with the number of positives with peak performance achieved using $40$ positives. Since during the sampling process of the contrastive loss, the number of negatives $n_\text{neg}$ is fixed to 100, we believe maintaining a good ratio between positives and negatives is critical for aligning the model to our proxy target. Since the optimal ratio could also be dependent on the data distribution, we hope our findings can motivate further studies on modifying the objective based on the characteristics of the available ICL examples.
%(Raymond: is this related to the number of databases? there could be a similar example in each database)

\subsubsection{Batch Size}

\begin{figure}[ht!]
    \centering
    \includegraphics[width=\linewidth]{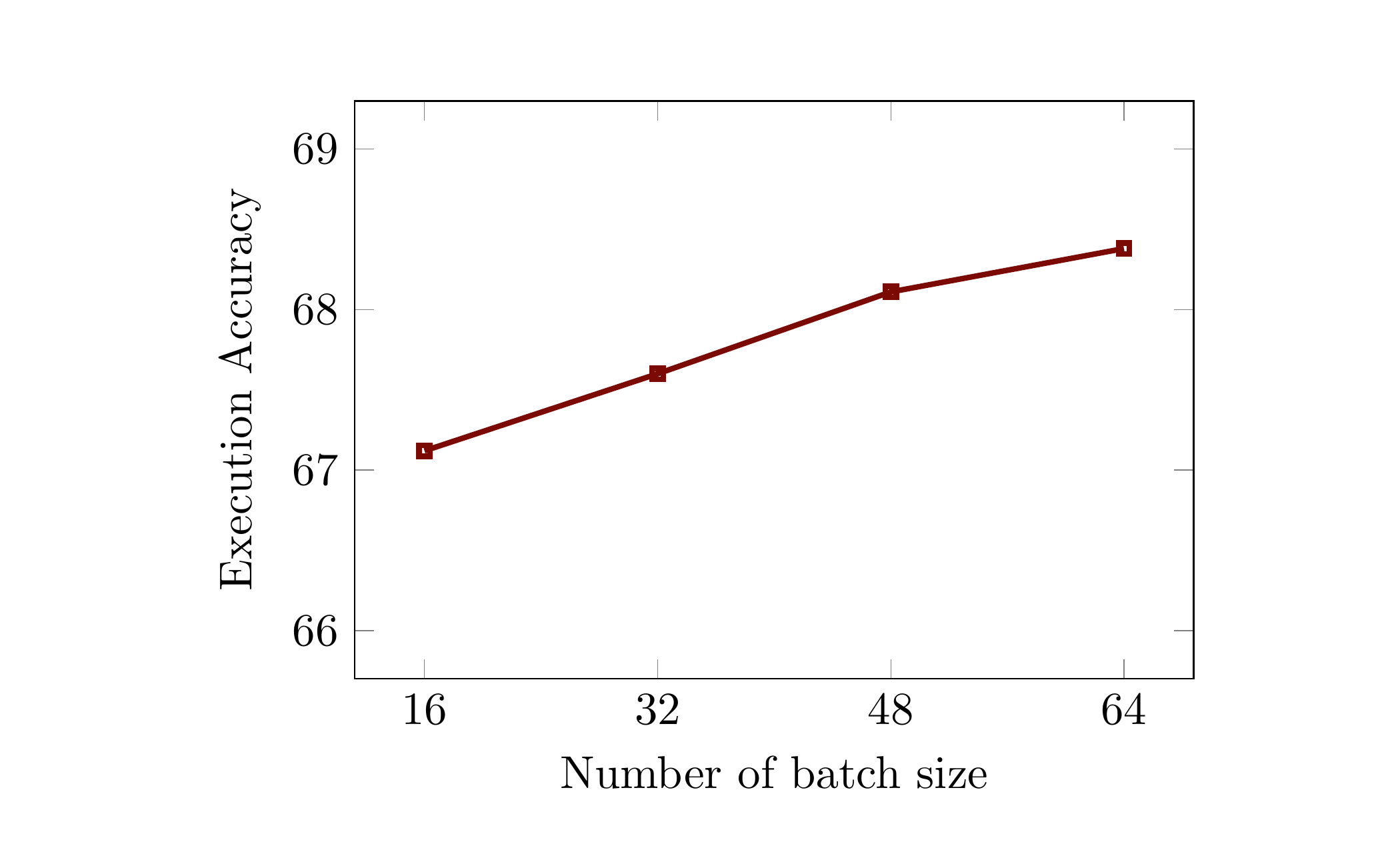}
    \caption{One-shot ICL execution accuracy on Spider where \ourmethod~is trained with different batch sizes.
    % (execution accuracy on development set of Spider dataset) of CodeLlama-13b where the prompt is retrieved by our trained \ourmethod~with a different batch size. The one-shot prompt is retrieved by the one with the largest cosine similarity.
    }%
    \label{fig:batchsize}
\end{figure}

%Raymond: not sure how to explain this, usually larger batch size = better performance.
\autoref{fig:batchsize} illustrates the effects of different batch sizes when training \ourmethod~with supervised contrastive loss. Consistent with the results from prior studies \citep{pmlr-v119-chen20j, khosla2020supervised}, we find that the task performance steadily improves as we increase the size of the mini-batch from 16 to 64.  
Since large batch sizes provide a more diverse set of negative examples, it enhances the model's ability to differentiate between subtle variations between examples, leading to superior task performance. We leave the experimentation of adapting more sophisticated metric learning techniques \citep{suárezdíaz2020tutorial} for demonstration retrieval as an exciting venue for follow-up works. \autoref{tab:diffresult} shows the performance of our \ourmethod~on examples with different difficulty levels. We find out that for a larger training batch size, \ourmethod~tends to perform better on easier data.

\begin{table}[h!]
    \centering
    \begin{tabular}{ccccc}
    \toprule
         Batch size& Easy & Medium & Hard & Extra  \\
    \midrule
       16  &  87.5& 66.14& 62.07& 43.37\\
       32  & 88.71& 67.49& 61.49& 42.77\\
       48  & 89.03 &68.76&62.57&42.16\\
       64  & 89.52&69.28&63.79&39.16\\
    \bottomrule
    \end{tabular}
    \caption{One-shot ICL execution accuracy for different classes on Spider development set of \ourmethod.}%
  \label{tab:diffresult}
\end{table}

\subsubsection{In-Domain Performance}
All prior experiments in this work are conducted in the out-of-domain (OOD) settings, where there are no schema overlaps between the query and reference set.  \yuxi{Nevertheless, when in-domain (ID) demonstrations are available, ID examples are usually better prompts in in-context learning for NL2SQL \citep{pourreza2024sqlencoder}. \citet{pourreza2024sqlencoder} trained an encoder-based retriever for in-domain demonstrations. To show the effectiveness of our \ourmethod~in the ID setting, we now study the effects of in-domain (ID) demonstrations where we directly retrieve ICL examples from the same schema in the evaluation set.} In the results reported in \autoref{tab:indomainresult}, we see that retrieving in-domain examples with \ourmethod~significantly outperforms the best ID baselines by $15.9$ accuracy points. This finding demonstrates that our trained \ourmethod~can effectively capture the nuances between examples under the same schema while emphasizing the advantages of having access to in-domain demonstrations.
Besides, we find that \ourmethod~(ID) achieves an astounding $13.4$ point improvement over \ourmethod~(OOD) retrieval model, indicating that \ourmethod~can generalize well when in-domain demonstrations are available.

\begin{table}[h!]
\small
 \centering
 \begin{tabular}{lc}\toprule
             & Exe Acc. $\uparrow$  \\\midrule
    OpenAI Embeddings$^\dag$& 69.1\\
    Cohere$^\dag$ & 69.3 \\
    SQL-Encoder$^\dag$ & 73.2\\
    %\ourmethod~(OOD)  & 68.4 \\
    \midrule
    \ourmethod~(ID)  & \textbf{86.6}\\%&\textbf{2.77}\\
    \bottomrule
 \end{tabular}
 \caption{Results on Spider dataset retrieved by development set data. $^\dag$Results are obtained from \citet{pourreza2024sqlencoder}}
 \label{tab:indomainresult}
\end{table}

% 3. hard negative mining

% 4. In domain table schema performance

% \paragraph{Cretiria to select positive examples}
% Execution Acc vs. PPL.

% \paragraph{Loss function}
% Triangle V.S. CE loss

% \paragraph{Attention weights}
% Qustion only/ DB schema

% \paragraph{In-domain performance (dev-dev)}

% \paragraph{Generalizability (another LM, e.g., Llama2 / or another benchmark, another prompt)}

%% file: fig/num_pos.tex
\begin{figure}[ht!] %插入图片

\centering %图片居中
\resizebox{\columnwidth}{!}{
\begin{tikzpicture}
\begin{axis}[
ybar,
legend pos=north west,
xlabel=Number of positive examples, %x坐标名
ylabel=Execution Accuracy, %y坐标名
ymin=65,
ymax=70,
enlargelimits=0.1,
width=10cm, height=5cm,
nodes near coords,
 nodes near coords align={vertical},
 nodes near coords style={font=\small},
%x tick label style={shift={(axis cs:0.5,0)},anchor=east,rotate=60,},
symbolic x coords={5,10,20,40,60,80},
]

    \addplot [ybar,fill=blue!40!white,draw=black, 
    postaction={pattern=north east lines} ] coordinates{
    %\addplot+[ybar,fill=RYB1] plot coordinates {
    (5,66.44)
    (10, 67.12)
    (20, 67.6)
    (40, 68.38)
    (60,66.92)
    (80, 67.02)
    };
    %\addlegendentry{Mean}
\end{axis}
\end{tikzpicture}
}
\caption{One-shot ICL results (execution accuracy on development set of Spider dataset) of CodeLlama-13b where the prompt is retrieved by our trained retriever with a different number of positive examples. The one-shot Prompt is retrieved by the one with the largest cosine similarity.} % 设置caption
\label{fig:num_pos}  % 设置用于reference的label
\end{figure}